\documentclass[runningheads]{llncs}
\usepackage{graphicx}

\usepackage{amsmath,amssymb,amsfonts}
\usepackage{algorithmic, algorithm}
\usepackage{adjustbox}
\usepackage{multirow}
\usepackage{multicol}
\usepackage{amssymb}

\def\etal{\emph{et al}}

\def\bmvaHangBox#1{
\begin{minipage}[t]{\textwidth}
\begin{tabbing} 
~\\[-\baselineskip] 
#1 
\end{tabbing}
\end{minipage}} 

\begin{document}
\title{Explaining How Deep Neural Networks \\ Forget by Deep Visualization}
%
%
\author{Giang Nguyen\inst{1} \and
Shuan Chen\inst{2} \and
Tae Joon Jun\inst{3} \and
Daeyoung Kim\inst{1}}
\authorrunning{G. Nguyen et al.}
%
\institute{School of Computing, KAIST, South Korea \\
\email{\{dexter.nguyen7,kimd\}@kaist.ac.kr}
\and
Department of Chemical and Biomolecular Engineering, KAIST, South Korea 
\email{shuankaist@kaist.ac.kr}\\
\and
Asan Institute for Life Sciences, Asan Medical Center, South Korea
\email{saigram89@gmail.com}\\}
\maketitle              
\begin{abstract}
Explaining the behaviors of deep neural networks, usually considered as black boxes, is critical especially when they are now being adopted over diverse aspects of human life. Taking the advantages of interpretable machine learning (interpretable ML), this paper proposes a novel tool called Catastrophic Forgetting Dissector (or CFD) to explain catastrophic forgetting in continual learning settings. We also introduce a new method called Critical Freezing based on the observations of our tool. Experiments on ResNet-50 articulate how catastrophic forgetting happens, particularly showing which components of this famous network are forgetting. Our new continual learning algorithm defeats various recent techniques by a significant margin, proving the capability of the investigation. Critical freezing not only attacks catastrophic forgetting but also exposes explainability.

\end{abstract}
\section{Introduction}
\label{sec:intro}

Regarding human evolution, life-long learning has been considered as one of the most crucial abilities, helping us develop more complicated skills throughout the lifetime. The idea of this learning strategy is hence deployed extensively by the deep learning community. Life-long learning (or continual learning) enables machine learning models to perceive new knowledge while simultaneously exposing backward-forward transfer, non-forgetting, or few-show learning \cite{ling2019unified}. While the aforementioned properties are the ultimate goals for life-long learning systems, catastrophic forgetting or semantic drift naturally occurs in deep neural networks in life-long learning settings because they are vastly optimized upon gradient descent algorithm \cite{goodfellow2013empirical}. 

\begin{figure*}
\begin{center}
\includegraphics[scale=0.3]{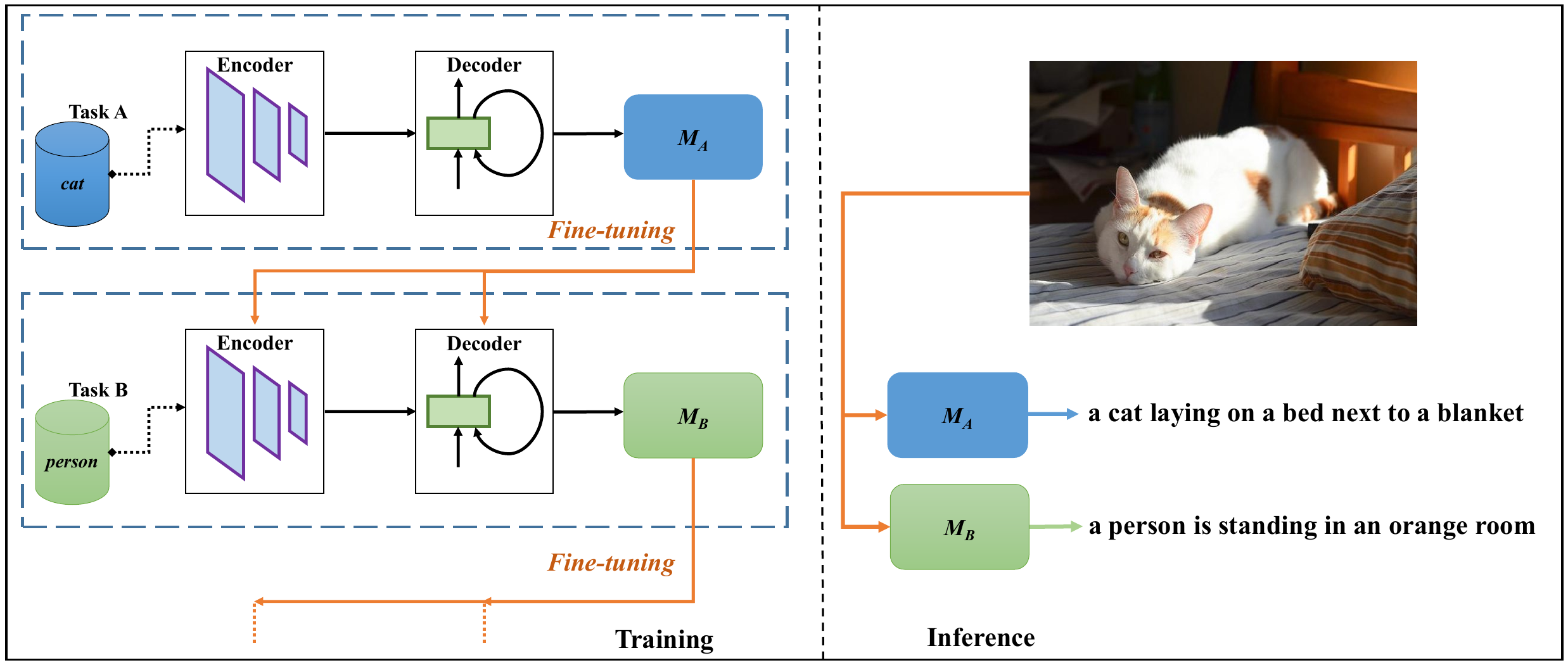}
\end{center}
   \caption{Catastrophic forgetting in continual learning settings.}
\label{fig:problem}
\end{figure*}

Catastrophic forgetting is defined as when we use a trained model on a given domain to address a new task, due to adapting to the new data samples, the model forgets what it learned before on the old domain. As shown in Fig. \ref{fig:problem}, if we use fine-tuning for continual learning, the knowledge acquired from the previous task will be eradicated and the inference becomes irrelevant. In fact, the description of the cat photo is shifted from ``\textit{a cat laying on a bed next to a blanket}'' to ``\textit{a person is standing in an orange room}''.

Although catastrophic forgetting is tough and undesirable, research to understand this problem is rare amongst the deep learning community. The interest in understanding or measuring catastrophic forgetting does not commensurate with the number of research to deal with this problem. Kemker \etal{} \cite{kemker2018measuring} develop new metrics to help compare continual learning techniques fairly and directly. Nguyen \etal{} \cite{nguyen2019toward} study which properties cause the hardness for the learning process. By modeling the chosen properties using task space, they can estimate how much a model forgets in a sequential learning scenario, shedding light on factors affecting the error rate on a task sequence. However, these research cannot show us what is being forgotten or which components are forgetting inside the model, but revealing what properties of tasks trigger catastrophic forgetting. By comparison, our work focuses on understanding which components of a network are volatile corresponding to a given sequence of tasks to articulate catastrophic forgetting.

More specifically, this research introduces a novel approach to elaborate catastrophic forgetting by visualizing hidden layers in class-incremental learning (considered as the hardest scenario in continual learning). In this learning paradigm, the use of previous data is prohibited and instances of the incoming tasks are unseen. We develop a tool named Catastrophic Forgetting Dissector (or \texttt{CFD}) which automates the dissection of catastrophic forgetting, exactly pointing out which components, in a model, are causing the forgetting. We formally adopt Intersection over Union (IoU), a popular evaluation metric in detection and segmentation tasks which essentially computes the overlapping ratio between two frames, to measure the forgetting degree of deep neural networks in this paper. The degree of forgetting is objectively measured after each class is added, thus giving us an intuition of how forgetting happens on a given part of the network.

A work from Kemker \etal{} \cite{kemker2018measuring} conducts experiments on state-of-the-art continual learning techniques that address catastrophic forgetting. It is demonstrated that the algorithms work, but only on weak constrains and unfair baselines, thus the forgetting problem is not fully solved and witnessed yet. They insist on the infeasibility of using toy datasets, such as MNIST \cite{kirkpatrick2017overcoming} or CIFAR \cite{zenke2017continual} in continual learning experiments. As a result, this work motivates us to choose Split MS-COCO \cite{nguyen2019contcap} to comprehensively measure the forgetting on deep neural networks. From the results of the dissection, we try to infer the plastic components in the network to preserve the accumulated knowledge. Critical freezing protects these components by simply keeping weights unchanged while training the new network. 

There are three main contributions in this work: 1) We propose a novel and pioneering method to analyze catastrophic forgetting in continual learning. 2) We introduce a new approach to mitigate catastrophic forgetting based on the findings. 3) Our extensive experiments demonstrate the efficacy of critical freezing and suggest recondite understanding about catastrophic forgetting. 

\section{Related Work}
\label{sec:relatedwork}
\paragraph{Feature Visualization}
Contemporary interpretability methods bring us advantages to understand the decision-making process of deep neural networks, ranging from visualizing saliency maps \cite{simonyan2013deep}\cite{dabkowski2017real} to transforming models into human-friendly structures \cite{che2016interpretable}. 
In \cite{zeiler2014visualizing}, deconvnets allow us to recognize which features are expected by a specific part of a network or what properties of image excite a chosen neuron the most.
In stark contrast to feeding an input image to diagnose, Yosinski \etal{} \cite{yosinski2015understanding} attempt to generate an image which maximizes the activation of a given neuron by gradient descent algorithm.
Visualizing the activation of a neuron or a layer in networks helps us categorize the specific role of each block, layer, or even a node. It has been proved that the earlier layers extract local features, such as edges or colors; while deeper layers are responsible for detecting globally distinctive characteristics. Prediction Difference Analysis (PDA) \cite{zintgraf2017visualizing}, even more specifically, highlights pixels that support or counteract a certain class, indicating which features are positive or negative to a prediction.  
However, these tools only provide the computer vision, leaving the conclusion for users. This manual process cannot ensure the quality of the observation when we may have hundreds or even thousands of attribution maps. \texttt{CFD} automatically detects the forgetting components in a network, entirely leveraging the generated activation difference heatmaps from PDA \cite{zintgraf2017visualizing}.  

\paragraph{Catastrophic Forgetting}
Many works have managed to address the forgetting problem in generative models \cite{zhai2019lifelong}, object detection \cite{shmelkov2017incremental}, semantic segmentation \cite{tasar2019incremental}, or captioning \cite{nguyen2019contcap}. Besides, fine-tuning is considered as a baseline in \cite{nguyen2019contcap}\cite{shmelkov2017incremental}\cite{tasar2019incremental}\cite{michieli2019incremental} to consider the superiority of the proposed techniques. Freezing either a few specific layers or a major part of a network is proposed in \cite{nguyen2019contcap}\cite{michieli2019incremental}, which reveals that just simply keeping some parameters unchanged can greatly help networks become robust against catastrophic forgetting. Learning without Forgetting (LwF) \cite{li2017learning} utilizes old models to generate pseudo data, which helps the new model reach a shared low-error region of problems. Also leveraging the old networks, knowledge distillation approaches \cite{michieli2019incremental}\cite{hou2018lifelong} are formally recognized to facilitate better generalization in life-long learning via a teacher-student learning strategy. Nevertheless, algorithms are inclined to rely on external factors (e.g., input or rehearsal data, objective functions \cite{li2017learning}\cite{michieli2019incremental}) while ignoring the question of why catastrophic forgetting internally happens. In this research, our algorithm is derived from catastrophic forgetting exploration.

\section{Approach}
\label{sec:approach}
Although research from \cite{nguyen2019toward} shows an interest in understanding catastrophic forgetting, they focus on how task properties influence the hardness of sequential learning. Hence, they are explaining based on the input data. \texttt{CFD} approaches the problem from an alternative perspective, trying to explain how forgetting happens over time based on the computer vision of models. In comparison, the ultimate goal of this tool is to figure out the most plastic conv layers or blocks in a network. Plasticity means a low degree of stiffness or being easy to change. Although a variety of continual learning techniques have been proposed to alleviate catastrophic forgetting, none of them takes advantage of the findings from Interpretable ML. Critical freezing is built on the top of \texttt{CFD}'s investigation to provide an interpretable and effective approach to deal with catastrophic forgetting. In the learning process, the optimal state of the old model is employed to initialize the new network. This way mimics the working mechanism of the human brain.

\subsection{Catastrophic Forgetting Dissector - CFD}
To dissect the model, we visualize the activation difference maps of hidden layers with respect to a specific unit (likely in the last layer) to understand the forgetting effect. We initially hypothesized that different features of the objects could be captured and visualized by particular feature maps in different layers. By looking into each response map in one conv block, we realize diverse features, such as eyes, face shape, car wheels, or background are isolatedly recognized by different channels, which avers our hypothesis. Unfortunately, considering each feature map manually by human eyes to study the forgetting is inefficient. To solve this issue, we compare the activation difference maps of feature maps with the ground truth segmentation to choose just one representative map in each conv block when the model has been trained on the first task. When new tasks arrive, the compare activation difference maps of two old and new models, in the same conv block, to measure the forgetting degree. In general, we do not seek for the answer that what features are being forgotten, but which conv blocks are forgetting.

\begin{figure*}
\begin{tabular}{cc}
\bmvaHangBox{\includegraphics[width=5.0cm]{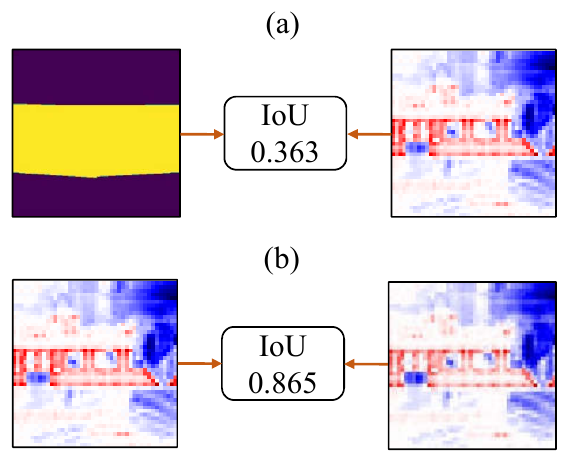}}&
\bmvaHangBox{\includegraphics[width=6.7cm]{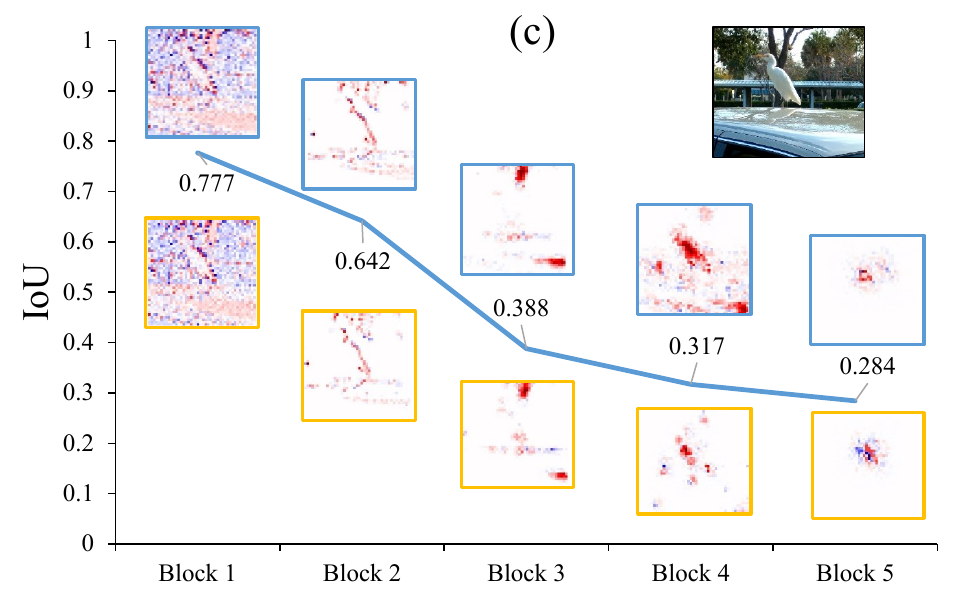}}\\
\end{tabular}
   \caption{(a) IoU value between the segmentation of a train and the positive features. (b) The IoU on the positive features of two representative maps. Red is evidence (or activate a unit )and blue is against (or decrease activation). (c) IoU graph and computer vision of a bird image. Yellow-bounding and blue-bounding boxes are from the old and new model respectively.}
\label{fig:IoU1}
\end{figure*}

The visualizations of the tool and the ground truth segmentation are simultaneously employed to infer the forgetting blocks. We assume that the semantic segmentation label of MS-COCO dataset \cite{lin2014microsoft} is what human eyes perceive. Next, we compare this segmentation with the computer vision of the model, particularly concentrating on positive evidence for a prediction to see how supportive features are disregarded.

The IoU value between the segmentation and evidence is calculated as shown in Fig. \ref{fig:IoU1} (a). On the right side of Fig. \ref{fig:IoU1} (a), we have an input image of a train, red dots advocate the fact that the output should be ``\textit{train}'' while blue ones contradict this prediction.   

Having the \textit{m-th} activation difference map ($FM$) in the \textit{l-th} block of a model \textit{M} and the ground truth segmentation \textit{GT}, the IoU is computed as:
\begin{equation}
\label{eq:1}
IoU_{M,GT} (l,m) =  \frac{FM(l,m) \cap GT }{FM(l,m) \cup GT}
\end{equation}

To select the activation difference map having the largest overlap with the ground truth in \textit{l-th} conv block, the representative map (RM) with the best IoU is $RM_{M,GT}$ :
\begin{equation}
\label{eq:2}
RM_{M,GT}  (l) = argmax_m (IoU_{M,GT} (l,m))
\end{equation}

To understand how the computer vision changes over the training process, we compare the $RM$s in the new model with the $RM$s of the old model shown in Fig. \ref{fig:IoU1} (b). We compute $IoU_{M_O,GT}$ by \eqref{eq:1} then achieve $RM_{M_O,GT}$ from \eqref{eq:2} ($RM_{M_O,GT}$ is the representative map of the old model). The forgetting effect of each trained model is measured by the IoUs between $RM_{M_O,GT}$ and activation difference maps of \textit{$M_N$} ($M_N$ is the new model):
\begin{equation}
\label{eq:3}
IoU_{M_N,M_O} (l,m) = \frac{F M(m) \cap RM_{M_O,GT} }{F M(m) \cup RM_{M_O,GT}}
\end{equation}

Similar to the method of finding out the best map fitting with ground truth, the map representing the best memory of the original activation difference map is denoted as $RM_{M_N,M_O}$ in \eqref{eq:4}. $RM_{M_N,M_O}$ is determined as: 
    
\begin{equation}
\label{eq:4}
RM_{M_N,M_O}  (l) = argmax_m (IoU_{M_N,M_O} (l,m))
\end{equation}

In the same block of both the old and new model, the role of a filter can be adjusted. For instance, the $50^{th}$ filter in the $2^{nd}$ block of the old model detects the eyes, but the same filter in the same block of the new model may consider the face. Hence, we should not make a comparison based on the index of a filter.

\begin{algorithm}[bt]
   \caption{CFD}
   \label{alg:cfd}
\begin{algorithmic}[1]
   \STATE {\bfseries Input:} Sample set $S$, segmentation ground truth $GT$, old model $M_O$, new model $M_N$, number of blocks $K$ 
   \STATE {\bfseries Output:} Forgetting conv block $\mathbb{F}$
   \STATE $i=0$
   \STATE $\L = \emptyset$
   \REPEAT
   \STATE $I = S[i]$
   \STATE $IoUs = \emptyset$
   \STATE $FM = PDA(I)$
   \FOR{$j=1$ {\bfseries to} $K$}
   \STATE $RM_{M_O,GT} \leftarrow \textit{FM with highest } {IoU_{M_O,GT}}$
   \STATE $RM_{M_N,M_O} \leftarrow \textit{FM with highest } {IoU_{M_N,M_O}}$
   \STATE \textbf{Append}($IoUs, max(IoU_{M_N,M_O})$)
   \ENDFOR
   \STATE $\boldsymbol{b} \leftarrow \textit{blocks with the highest drop in IoUs}$ 
   \STATE \textbf{Append}(\L, $\boldsymbol{b}$)
   \STATE $i = i+1$
   \UNTIL{$i =  \textbf{size}($S$)$}
   \STATE $\mathbb{F}$ $\leftarrow$ \textit{Most frequent block in} \L
\end{algorithmic}
\end{algorithm}
    
The workflow of \texttt{CFD} is given by Algorithm \ref{alg:cfd}. The sample set $S$ is particularized in Section \ref{sec:exp}, $GT$ is the segmentation ground-truth from MS-COCO dataset, $M_{O}$ and $M_{N}$ are the old and new model for comparison respectively, and $K$ is the number of the conv blocks in the network ($K=5$ with ResNets). $\L$ is a list containing the most forgetting conv block with respect to all the images in $S$. By inputting an image $I$ from $S$, we get the visualization of maps ($FM$) over $M_{O}$ and $M_{N}$ by PDA. However, we need to pick the representative activation difference map (RM) amongst thousands of maps in a conv block.

To choose the representative map $RM$ of the $j^{th}$ conv block in a model, we define representative map $RM$ to be the map having the largest overlap with the $RM$ of the previous model ($RM_{prev}$ in short) for the same $j^{th}$ block. Particularly, the $RM_{prev}$ of $M_{O}$ is the ground truth because $M_{O}$ is the starting model, and $RM_{M_O,GT}$ is the map at $j^{th}$ block of $M_{O}$. Likewise, $RM_{prev}$ of $M_{N}$ at $j^{th}$ block is $RM_{M_O,GT}$, and we obtain $RM_{M_N,M_O}$ by comparing maps of $M_{N}$ and $RM_{M_O,GT}$. The IoU values between $RM_{M_O,GT}$ and $RM_{M_N,M_O}$ are calculated at each conv block and appended to a list $IoUs$. After calculating IoU drops through the ResNet blocks and denote the block giving the highest drop as $b$, we can put $b$ as the block where the most substantial forgetting happens into a list $\L$, tested on the input image $I$. Finally, we generalize on all images of $S$ to return the most forgetting component $\mathbb{F}$. 

In the juxtaposition of the old and new model, the IoUs are visually drawn to provide a bird-eye view of the forgetting trend shown in Fig. \ref{fig:IoU1} (c). We may argue that the conv block having the lowest IoU (block 5) is the victim of catastrophic forgetting. However, this assumption is not asserted because of the error accumulation in deep neural networks. The computer vision of deeper blocks is directly attributed to earlier conv blocks. Once forgetting occurs in the first conv block, it will be propagated throughout the entire network. 

We propose to leverage IoU slopes to find the weakest block $b$. At a point, if the IoU drops significantly compared to the previous value, it should be the sign of catastrophic forgetting. In Fig. \ref{fig:IoU1} (c), a plummet of the IoU value is seen between block 2 and block 3 (0.642 to 0.388). The first thought appearing in our mind was that the $3^{rd}$ is forgetting most catastrophically. It is true but we need to regard the fact that the worst map in block 3 is a result of block 2 and block 1. This finding is the foundation for our technique to prevent catastrophic forgetting. 

\subsection{Critical Freezing}
\begin{algorithm}[hbt!]
   \caption{Critical Freezing}
   \label{alg:cf}
\begin{algorithmic}[1]
   \STATE {\bfseries Input:} Sample set $S$, segmentation ground truth $GT$, old model $M_O$, new model $M_N$, number of blocks $K$
   \STATE {\bfseries Output:} optimal state $\theta^{*}$ 
   \STATE $M_N$ $\leftarrow$ $M_O$          // initialize new model by old parameters
   \STATE $F$ $\leftarrow$ \texttt{CFD}($S$, $GT$, $M_O$, $M_N$, $K$)
   
   \FOR{$i=1$ {\bfseries to} $F-1$} 
   \STATE \textbf{grad}($M_N[i]$, $False$) // freeze the block $i^{th}$
   \ENDFOR
   
   \STATE $\theta^{*}$ $\leftarrow$ $\underset{\hat{\theta}}{argmin}(- \sum_{i=1}^{\mathbb{V}} Y_{k}^{i} \log \hat{Y}_{k}^i)$
\end{algorithmic}
\end{algorithm}

Fine-tuning techniques play a pivotal role in training deep networks if data distribution evolves. Regarding a pre-trained model, the feature extractor which captures global information is carefully protected in adaptation. The output layer may be superseded, or the learning rate should be tweaked to a tiny number. Another dominant approach is to freeze the weights of the early layers. They are all effective yet ambiguous because we cannot ensure freezing which layers will give the best result. Using the investigation from \texttt{CFD}, we freeze the precursors of the most plastic conv block in a deep neural network. If a network has $K$ conv blocks, and we find the $F^{th}$ convolutional block broken, then we try to freeze earlier blocks than the $F^{th}$ block. If updating the fragile components are necessary, a learning rate on those blocks should be thoroughly calibrated. The procedure of critical freezing is shown in Algorithm \ref{alg:cf}. The objective function is the standard cross-entropy loss for image captioning, $\mathbb{V}$ is the size of the vocabulary, $Y_{k}^{i}$ is the ground truth, and $\hat{Y}_{k}^i$ is the prediction.

\section{Experiments}
\label{sec:exp}
We use a dataset called Split MS-COCO from \cite{nguyen2019contcap} to reproduce catastrophic forgetting in image captioning task with incremental learning schemes. The dataset contains over 47k images for training and over 23k images for validation and testing. Regarding the incremental learning setup, a new class is introduced at each time step. Initially, we train with 19 classes to acquire the base model, followed by adding 5 classes sequentially. The captioning model is divided into an encoder and a decoder, in which the encoder is the ResNet-50, and the decoder includes an embedding layer, a single-layer LSTM, and a fully-connected layer producing a word at a time step. As \texttt{CFD} works on a single image, running multiple times on different and diverse input images is needed, helping us to generalize the observation of forgetting.

We choose a sample set $S$ (\textit{bicycle, car, motorcycle, airplane, bus, train, bird, cat, dog, horse, sheep, and cow}) from 19 trained classes. The results of \texttt{CFD} reinforce the fact that the $3^{rd}$ conv block of ResNet is the most plastic component of this famous conv net given the learning sequence.
To evaluate critical freezing, we perform fine-tuning and various schemes of freezing. In fine-tuning, the old model initializes the new model, and training is done by minimizing the loss on the new task. As the network contains two parts, encoder and decoder, we freeze them separately to specify the best freezing strategy. In addition, we choose various combinations of blocks to be frozen besides a famous baseline in continual learning experiments called LwF \cite{li2017learning}. Two knowledge distillation techniques from \cite{michieli2019incremental} are also taken into comparison. The traditional scores for image captioning are considered in evaluating the superiority of critical freezing over the baselines. $BLEU4$ and $ROUGE_L$ are essentially word-overlap based metrics, while CIDEr and SPICE are more trustworthy because they give more weights on significant terms, such as verbs or nouns. Therefore, we will give discussion only on CIDEr score for conciseness.   

\begin{figure*}[ht!]
\begin{center}
\includegraphics[scale=0.85]{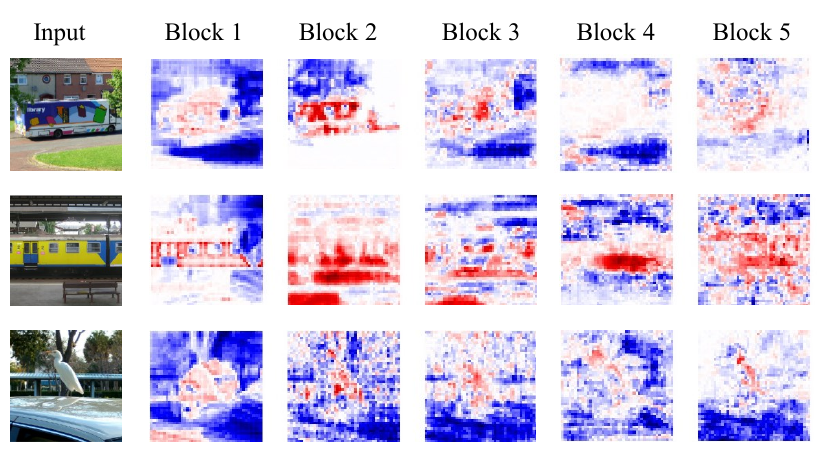}
\end{center}
   \caption{Activation difference maps from convolutional blocks of ResNet-50.}
\label{fig:Visualizing}
\end{figure*}

\begin{figure*}[bt]
\begin{tabular}{cc}
\bmvaHangBox{\includegraphics[width=6.5cm, height=5.0cm]{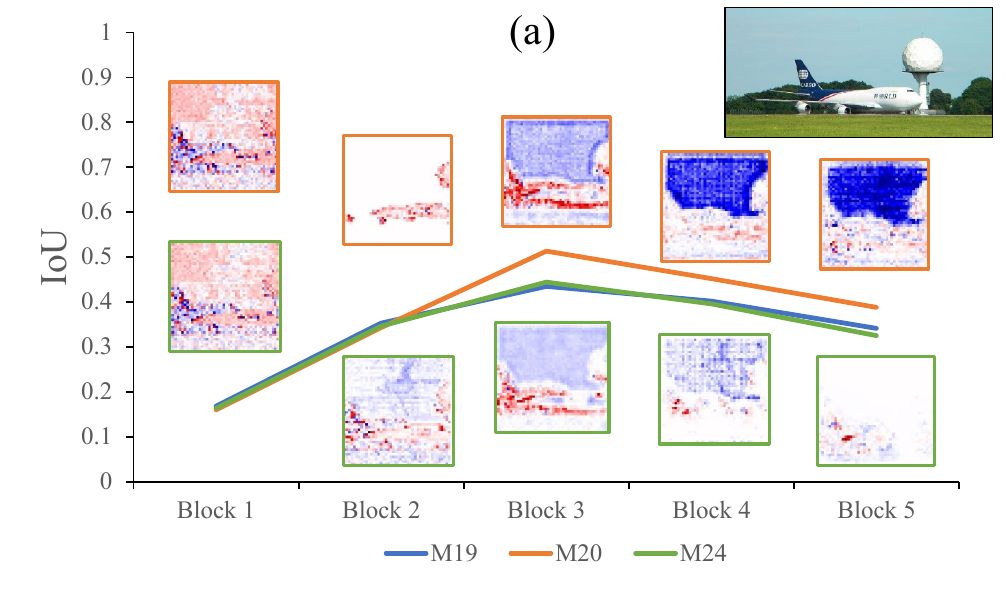}}&
\bmvaHangBox{\includegraphics[width=6.5cm, height=5.0cm]{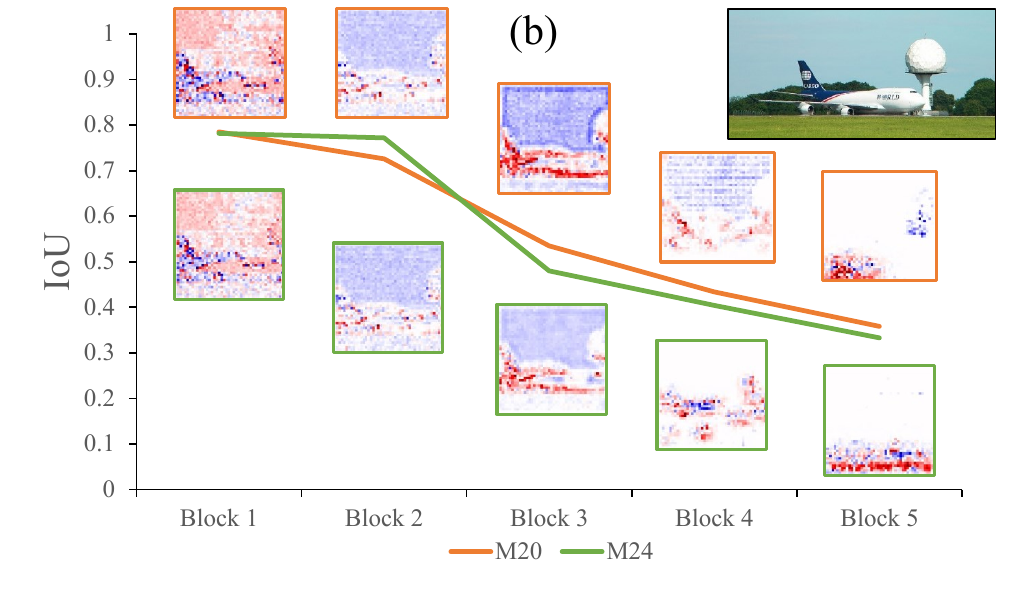}}\\
\end{tabular}
   \caption{(a) $IoU_{M,GT}$ of $M_{19}$, $M_{20}$ and $M_{24}$ comparing with GT. (b) $IoU_{MT,MO}$ of model $M_{20}$ and $M_{24}$ comparing with $M_{19}$.}
\label{fig:IoU_GT}
\end{figure*}

After adding a new class, we obtain $M_{20}$, and $M_{n}$ is the model when a total number of $n$ classes are witnessed. In Fig. \ref{fig:Visualizing}, the visualized results show that the first and second blocks of ResNet-50 can overall capture the outline of objects. Computer vision turns to represent more detailed features from the objects and other background features to determine the class of the input image in the deeper blocks. Also, the IoUs of different blocks in models, comparing with ground truth, are calculated by \eqref{eq:1} and \eqref{eq:2}. The results reveal that although different models show the performance of the classification inconsistently, the $IoU_{M_n,GT}$ ($n>19$) values are roughly similar at all the blocks, which implies that no matter the how good the performance is, the level of matching between maps of each model and the human vision is preserved (Fig. \ref{fig:IoU_GT} (a)).

To measure how much the forgetting occurs in the ResNet, we compute IoUs by \eqref{eq:3} and \eqref{eq:4}. It is clearly shown in Fig. \ref{fig:IoU_GT} (b) that the IoUs between $M_{20}$ and $M_{19}$ are roughly equal to the corresponding figure for $M_{24}$ and $M_{19}$ in every block. For $M_{20}$, the $IoU_{M_{20},M_{19}}$ is always around 0.8 at the first block, showing that a marginal forgetting happens here. The $IoU_{M_{20},M_{19}}$ starts to drop along the blocks because the later maps are constructed by the previous maps. The forgetting effect persists and does not show which block is forgetting the most. 

For $M_{24}$, the first block of the model still gets a high IoU comparing with $M_{19}$ and the values decrease from the second block.
Unlike the constantly decreasing trend seen in $M_{20}$, the decreasing rate of $IoU_{M_{24}, M_{19}}$ fluctuates through the blocks and a severe drop at block 3 is observed in every testing input, suggesting that the forgetting effect might happen the most in this block. Iterating this procedure on all images of the set $S$ reinforces that the worst forgetting happens at block 3.

\begin{table*}[hbt]
\caption{Performance when 5 classes arrive sequentially on past tasks and newly added tasks.}
\begin{center}
\begin{adjustbox}{width=1.0\textwidth,center=\textwidth}
\begin{tabular}{|l|c|c|c|c|c|c|c|c|}
\hline
\multirow{2}{*}{} & \multicolumn{4}{c|}{$Past-task$}                         & 
\multicolumn{4}{c|}{$New-task$}              \\ \cline{2-9} 
                           & BLEU4         & ROUGE\_L       & CIDEr     & SPICE                         & BLEU4        & ROUGE\_L       & CIDEr     & SPICE     \\ \hline
\textbf{Fine-tuning}& 4.2                   & 32.3          & 4.9   & 1.9               & 10.1                    & 39.6         & 17.9      & 5.9    \\ \hline
\textbf{Encoder-Freezing}& 3.8                  & 31.5          & 5.2    & 2.0            & 8.8                    & 38.4          & 15.1      & 5.1    \\ \hline
\textbf{Decoder-Freezing}& 5.2                   & 33.3          & 6.8   & 2.2                                      & 10.6         & 39.8          & 18.5   & 5.5   \\ \hline
\textbf{L1-Freezing}& 5.2       & 33.5         & 7.9    & 2.4         & 12.0      & 41.1          & 23.4      & 6.4    \\ \hline
\textbf{L2-Freezing}& 5.8                   & \textbf{33.6}         & 7.4    & 2.1            & 12.0                   & \textbf{41.2}          & 24.0      & 6.5    \\ \hline
\textbf{L3-Freezing}& 4.5                   & 32.2         & 6.8    & 2.1          & 10.4                   & 39.6          & 20.7      & 6.0    \\ \hline
\textbf{L4-Freezing}& 5.0                   & 32.6         & 7.6    & 2.2            & 11.4                   & 40.0          & 23.3      & 6.2    \\ \hline
\textbf{L5-Freezing}& 5.4                   & 33.0         & 7.4    & 2.3            & 12.1                   & 40.4          & 23.3      & 6.2    \\ \hline
\textbf{LwF\cite{li2017learning}}& \textbf{6.4}                  & 33.2          & \textbf{9.7}   & 2.6                    & 10.6        & 38.9          & 16.1     & 5.3     \\ \hline
\textbf{KD1 \cite{michieli2019incremental}}& 4.7                  & 33.3          & 6.7   & 2.1                    & 11.4        & 40.4          & 20.8     & 6.2     \\ \hline
\textbf{KD2 \cite{michieli2019incremental}}& 3.9                  & 32.4     & 5.6      & 2.0     & 10.6   & 39.8          & 18.6     & 5.9     \\ \hline
\textbf{Critical Freezing}& 5.6                  & 33.2               & 9.5   & \textbf{2.8}                   & \textbf{12.2}                    & 40.7          & \textbf{26.1}       & \textbf{6.9}   \\ \hline
\end{tabular}
\end{adjustbox}
\end{center}
\label{tab:my-table}
\end{table*}

While two naive approaches of freezing in \cite{nguyen2019contcap} are also implemented, we devise critical freezing based on findings, which only freezes critical conv blocks. As shown in Table. \ref{tab:my-table}, precisely freezing helps to learn on both the new and old tasks much more effectively. Our freezing scheme outperforms the other approaches on new tasks by a large margin (26.1 CIDEr) while achieving comparable performance with LwF \cite{li2017learning} on past tasks (9.5 CIDEr) although LwF \cite{li2017learning} is far more complicated. Knowledge distillation on intermediate feature space (KD1) and output layer (KD2) \cite{michieli2019incremental} claims 20.8 and 18.6 CIDEr respectively. We argue that the frozen blocks contain global information derived from past tasks, which is valuable and should be accumulated during lifetime rather than changing. Fine-tuning optimizes the loss on the new task without any guidance; as a result, the model may not fall into the low-error regions of tasks. We try to freeze each conv block of the ResNet to fortify the hypothesis that properly freezing is really better than ambiguous freezing in fine-tuning schemes. Hence, critical freezing exerts a promising influence on fine-tuning techniques in deep learning.     

\section{Conclusion and Future Work}
As the presence of catastrophic forgetting hinders the life-long learning, understanding how this phenomenon happens in computer vision is imperative. We introduce \texttt{CFD} to grasp catastrophic forgetting. The investigation of our tool unearths the mystical question about catastrophic forgetting. 

From knowing where the forgetting issue is coming from, a new technique has been proposed focusing on plastic components of a model to moderate the information loss. The experiments illustrate the superiority of critical freezing over various freezing schemes and existing techniques. To the best of our knowledge, no work has been done for mitigating catastrophic forgetting under the supervision of Interpretable ML. By knowing which regions are needed to be kept intact, not only could the performance on the old task be largely improved, but the new task is also more conquerable. Ultimately, critical freezing could benefit a variety of fine-tuning schemes and continual learning approaches.

There are future works following our paper. Scaling this work for other tasks (e.g. \cite{tran2021simple}) and deep networks can better validate the proposed continual learning algorithm's feasibility. We believe that our work results are just a starting point for a new direction to address catastrophic forgetting by interpretable methods completely. The experiments on conv blocks are coarse, which could be further refined by using conv layers, which will likely give much better results. We are choosing PDA \cite{zintgraf2017visualizing} because the maps are clear; however, we can also adopt other advanced methods, such as using attribution or saliency to analyze the forgetting problem better.

\section*{ACKNOWLEDGMENT}
This work was supported by Korea-EU Joint Research Support Project through the Ministry of Science and ICT (MSIT) and National Research Foundation of Korea (NRF-2016K1A3A7A0395205414), and the Technology Innovation Program (or Industrial Strategic Technology development Program, 2000682, Development of Automated Driving Systems and Evaluation) funded by the Ministry of Trade, Industry and Energy (MOTIE, Korea).

%
%
%
%

\newpage
\bibliography{egbib}

\bibliographystyle{splncs04}

\end{document}